\documentclass{article}

\usepackage[preprint,nonatbib]{neurips_2024}

\usepackage{microtype}
\usepackage{graphicx}

\usepackage{booktabs} 
\usepackage{amsmath}
\usepackage{enumerate}[nolistsep]
\usepackage[shortlabels]{enumitem}

\usepackage{multirow}
\usepackage{algorithm}
\usepackage[noend]{algpseudocode}
\usepackage{threeparttable}
\usepackage{subcaption}
\usepackage{xurl}
\usepackage[numbers]{natbib}
\usepackage{hyperref}

\usepackage{amsmath}
\usepackage{amssymb}
\usepackage{mathtools}
\usepackage{amsthm}
\usepackage{multirow}

\usepackage{array}
\usepackage[capitalize,noabbrev]{cleveref}

\theoremstyle{plain}
\newtheorem{theorem}{Theorem}[section]
\newtheorem{proposition}[theorem]{Proposition}

\theoremstyle{definition}

\theoremstyle{remark}

\usepackage[textsize=tiny]{todonotes}
\title{Debiasing Kernel-Based Generative Models}

\author{%
  Tian Qin\\
  Math department\\
  Lehigh University\\
  Bethlehem, PA 18015 \\
  \texttt{tiq218@lehigh.edu} \\
  \And
  Wei-Min Huang\\
  Math department\\
  Lehigh University\\
  Bethlehem, PA 18015 \\
  \texttt{wh02@lehigh.edu} \\
}

\begin{document}

\maketitle

\begin{abstract}
We propose a novel two-stage framework of generative models named Debiasing Kernel-Based Generative Models (DKGM) with the insights from kernel density estimation (KDE) \cite{Parzen,Rosenblatt} and stochastic approximation \cite{Stochastic_Approximation}. In the first stage of DKGM, we employ KDE to bypass the obstacles in estimating the density of data without losing too much image quality. One characteristic of KDE is oversmoothing, which makes the generated image blurry. Therefore, in the second stage, we formulate the process of reducing the blurriness of images as a statistical debiasing problem and develop a novel iterative algorithm to improve image quality, which is inspired by the stochastic approximation \cite{Stochastic_Approximation}. Extensive experiments illustrate that the image quality of DKGM on CIFAR10 is comparable to state-of-the-art models such as diffusion models \cite{DDPM} and GAN models\cite{SNGAN,SNGAN-DDLS,StyleGAN2+ADA}. The performance of DKGM on CelebA 128$\times$128 and LSUN (Church) 128$\times$128 is also competitive. We conduct extra experiments to exploit how the bandwidth in KDE affects the sample diversity and debiasing effect of DKGM. The connections between DKGM and score-based models \cite{score-based} are also discussed. 
\end{abstract}

\section{Introduction}
\label{Introduction}

Generative models have been used to generate high-quality samples in many fields of machine learning. More specifically,  Generative adversarial networks (GANs), autoregressive models, and diffusion models have synthesized high-fidelity images and realistic speech and video\citep{GAN,autoregressive,DDPM,gan_image,video_network,menick2018generating,audio,video_diffusion}. Moreover, diffusion models gained popularity in the field of image generation. Most of the generative models share the spirit of sampling new data from pure noise, and the generated data should be consistent with the observed data distribution as much as possible. For example, VAE \cite{VAE}, GAN \cite{GAN} and diffusion models \cite{DDPM} typically estimate the density of the image distribution $p(\mathbf{x})$ for generation. As a result, VAE suffers from the restricted structure of latent space;  GAN falls ill with model collapse due to the subtleness of objective function and diffusion model is slow in generating new images with dedicated backward process. On the other hand, modern generative models always involve two structures. For instance, VAE requires encoder and decoder; GAN needs generator and discriminator and diffusion models has forward and backward processes.

In this paper, we will propose a novel two-stage (asymmetric structure) model to bypass the obstacles in estimating the density of data without losing too much image quality. More specifically, the main contributions of our work can be summarized as followings:
\begin{enumerate} 
    \item We first formulate the generation task as a problem of kernel density estimation, which is the first stage of our model.
    \item As to the second stage, we propose a novel debiasing algorithm to enhance the quality of KDE sampled image. The idea is inspired by the stochastic approximation proposed in \cite{Stochastic_Approximation}. The entire two-stage model is named Debiasing Kernel Based Generative Model(DKGM).
    \item  The implementation of DKGM involves two networks, one is used to generate initial new images by kernel density sampling, the other is to improve the image quality. The experiments demonstrate that DKGM is comparable to many baseline methods in the sense of FID  and inception score. The ablation studies also show that the stage 2 of DKGM improves the sample quality (sharpness)\cite{WAE} in general. 
\end{enumerate}

 The remaining sections of this paper are organized as following schema.  In section \ref{preliminary}, we provided some preliminaries about KDE, stochastic approximation and  theoretical results. Based on analysis in section \ref{preliminary}, we introduced the two-stage model DKGM in section \ref{Iterative Debiasing}.  The comparisons between DKGM and other modern generative models in benchmark datasets are illustrated  in section \ref{Experiments}. In addition, we constructed a numerical example to show the effect of the proposed debiasing algorithm in section \ref{toy example}.  Section \ref{discussion} discusses  limitations of DGKM and suggests some future directions.
 



\section{Preliminary}
\label{preliminary}
\subsection{Kernel density estimation (KDE)}

For simplicity, we only consider the one-dimension case. Given $X_{1},...,X_{m}$ be i.i.d random variables with a continuous density function $f$, \cite{Parzen} and \cite{Rosenblatt} proposed kernel density  estimate $p_{m}(x)$ for estimating $p(x)$ at a fixed point $x\in \mathbb{R}$,
\begin{equation}
\label{kernel estimate}
    p_{m}(x)=\frac{1}{mb_{m}}\sum_{k=1}^{m}K\left( \frac{x-X_{k}}{b_{m}}\right)
\end{equation}
where  $K$ is an appropriate kernel function such that $\int K(x)dx=1$ and the positive number $b_{m}$ is called bandwidth such that $b_{m}\to 0, mb_{m}\to \infty$ as $m\to \infty$. By convention, literature about VAE \cite{VAE}, GAN \cite{GAN} and Diffusion models \cite{DDPM} typically estimates data distribution $p(\mathbf{x})$ for generation. However the generation with KDE is straightforward as we can essentially generate new data without deriving or estimating the data density. More specifically, if we let the kernel be Guassian distribution, sampling from the corresponding kernel density  estimate $p_{m}(x)$ involves two steps:

\begin{enumerate}
    \item Randomly pick  one data point $x_{i}$
 from the data $\{x_1,x_2,...,x_{n}\}$ included in the KDE
 \item Given point $x_{i}$, draw a value  from the Gaussian $N(x_{i},b_{m}^{2})$ centered at $x_{i}$ and of standard deviation $b_{m}$
 (the bandwidth)
\end{enumerate}

Approximating the true density function $p$ with $p_{n}(x)$, we have the following inequality of empirical likelihood $\text{log}(x)$:
\begin{equation}
    \label{ineq KDE}
    \begin{split}
      \text{log}p(x)\approx \text{log}p_{m}(x)&=\text{log}\left[\frac{1}{mb_{m}}\sum_{k=1}^{m}K\left( \frac{x-X_{k}}{b_{m}}\right) \right]  \\
      &\geq \sum_{k=1}^{m}\text{log}\bigg[\frac{1}{mb_{m}}K\left( \frac{x-X_{k}}{b_{m}}\right) \bigg] 
    \end{split}
\end{equation}

For a given datapoint $x$ and kernel function $K$, the smaller the distance between $x$ and KDE generated sample $X_{k}$ is, the larger the RHS of the inequality \eqref{ineq KDE} will be. To this sense, if we can model the a network generating the KDE samples around the original input $x$, the empirical likelihood  $\text{log}p(x)$ will be improved. Inequality \eqref{ineq KDE} inspires us to develop the kernel based generation model in section \ref{kernel generative model}. Next, we will introduce the concept of the bias of data restoration to  enhance the quality of KDE samples. 
\subsection{Bias of data restoration}
\label{Bias reduction}
Let $\mathbf{x}$\footnote{We will use bold letter $\mathbf{x}$ to denote the high-dimension data point.} be the original data and $\hat{\mathbf{x}}$ is the restored data from certain networks. If we view $\hat{\mathbf{x}}$ as an estimator of true data point,  it is natural to define the bias of restored data $\hat{\mathbf{x}}$ as following:
\begin{equation}
    \label{bias initial}
    \text{Bias}(\hat{\mathbf{x}})=\mathbf{x}-\mathbb{E}[\hat{\mathbf{x}}|\mathbf{x}]
\end{equation}
where the expectation is taken over all possible restored $\hat{\mathbf{x}}$ given true data is $\mathbf{x}$. 

This is an analogue of the mean bias of an estimator in statistic inference \cite{casella2002statistical}.  As one of the most important and mostly used concepts of statistical inference, unbiasedness of an estimator ensures that the target estimator estimates the unknown parameter $\theta$ properly in the sense of average. 

If the bias term in \eqref{bias initial} is negligible, we can say the restored data $\hat{\mathbf{x}}$ is satisfactory. Otherwise, it's natural to substract the estimated bias from initial restored data to obtain less biased data. However, the exact formula of bias in \eqref{bias initial} in unavailable in most cases as the derivation of the expectation is complicated in real world applications.  To tackle with this issue, for the initially reconstructed data $\hat{\mathbf{x}}$, we consider the following equation 
\begin{equation}
    \label{stationary eq}
    \hat{\mathbf{x}}=\mathbb{E}[\Tilde{\mathbf{x}}|\mathbf{x}^{*}]
\end{equation}
where $\Tilde{\mathbf{x}}$ is the version of restored data given input is $\mathbf{x}^{*}$ and the conditional expectation is taken over all possible restoration data $\Tilde{\mathbf{x}}$  given input data is $\mathbf{x}^{*}$. In other words, we want to solve for $\mathbf{x}^{*}$ which satisfies equation \eqref{stationary eq} and we claim that the solution $\mathbf{x}^{*}$ will be approximately unbiased with respect to the true data $\mathbf{x}$ under certain conditions, i.e $E[\mathbf{x}^{*}|\mathbf{x}]\approx \mathbf{x}$. Given $\hat{\mathbf{x}}$, we assume that the  restoration process is random\footnote{This assumption only aims to facilitate the theoretical derivation. In practice, the restoration stage (second stage) of DKGM doesn't have to be random as we model the conditional expectation by networks directly. } and the conditional expectation can be represented by a differentiable bijective map $H: \mathcal{X}\to \mathcal{X}$. For example, we have  $ \mathbb{E}[\hat{\mathbf{x}}|\mathbf{x}]=H(\mathbf{x})$ and   $\mathbb{E}[\Tilde{\mathbf{x}}|\mathbf{x}^{*}]=H(\mathbf{x}^{*})$. The bijectivity of map $H$ assures the one-to-one corresponding between $\hat{\mathbf{x}}$ and $\mathbf{x}^{*}$. That is to say, for each $\hat{\mathbf{x}}$ we can find a unique $\mathbf{x}^{*}$ and if $\hat{\mathbf{x}}$ is random, $\mathbf{x}^{*}$ will be random as well  .  In many real cases, we have
 $\mathcal{X}=\mathbb{R}^{p}$.  To illustrate the core idea, we assume $p=1$ in the following derivation.

Given true data $\mathbf{x}$, take the conditional expectation on both side of \eqref{stationary eq} , and we obtain
\begin{equation}
\label{expectation sta eq}
    \mathbb{E}[\hat{\mathbf{x}}|\mathbf{x}]= \mathbb{E}[H(\mathbf{x}^{*})|\mathbf{x}]
\end{equation}

Applying the Taylor expansion of $H(\mathbf{x}^{*})$ around $\mathbb{E}[\mathbf{x}^{*}|\mathbf{x}]$, we have
\begin{equation}
\label{taylor}
    \begin{split}
        H(\mathbf{x}^{*}) &=H(\mathbb{E}[\mathbf{x}^{*}|\mathbf{x}])+ H'(\mathbb{E}[\mathbf{x}^{*}|\mathbf{x}])\cdot (\mathbf{x}^{*}-\mathbb{E}[\mathbf{x}^{*}|\mathbf{x}])\\
        &+\frac{1}{2}H''( \mathbb{E}[\mathbf{x}^{*}|\mathbf{x}] )\cdot ( \mathbf{x}^{*}-\mathbb{E}[\mathbf{x}^{*}|\mathbf{x}] )^{2}\\
        &+O(|
        \mathbf{x}^{*}-\mathbb{E}[\mathbf{x}^{*}|\mathbf{x}]|^3)
    \end{split}
\end{equation}
If $H$ is nearly linear,i.e. $H'' \approx 0$,  plugging identity \eqref{taylor} into equation \eqref{expectation sta eq} gives us the following approximation
\[
H(\mathbf{x})\approx H(\mathbb{E}[\mathbf{x}^{*}|\mathbf{x}])
\]
The bijectivity of $H$ implies that the solution $\mathbf{x}^{*}$ of \eqref{stationary eq} is approximately unbiased, i.e. $\mathbb{E}[\mathbf{x}^{*}|\mathbf{x}] \approx \mathbf{x}$.  It can be viewed as a debiased version of initial restoration $\hat{\mathbf{x}}$. The  analysis above can be generalized to vector-valued function $H$ and works for the cases when $p>1$. Consequently, the bias reduction  boils down to solving the inverse problem in \eqref{stationary eq}, which can be approached by the stochastic approximation method invented by \cite{Stochastic_Approximation}. 
\subsection{Stochastic approximation}
\label{Stochastic approximation}

In one dimensional case, \cite{Stochastic_Approximation} proposed  to solve the equation $M(x)=\alpha$ where $M(x)=\int_{-\infty}^{\infty}ydF(y|x)$ by a stochastic sequence generated by following iteration:
\begin{equation}
\label{SA iteration}
   x_{k+1}-x_k=a_k(\alpha-y_k) 
\end{equation}
where $y_{k}$ is a realization of the random variable which has the distribution of $F(y|x_{k})$.

 Under certain regularity conditions, \cite{Stochastic_Approximation} shown that if sequence $a_{k}$ satisfies $\sum a_{k}=\infty$, and $\sum a^{2}_{k}<\infty$ then the iterated sequence of $\{x_{k}\}$ will converges to the root $\theta$ , which satisfies the original equation $M(\theta)=\alpha$, in $L^{2}$ (also in probability). \cite{SA_prob1}    proved a stronger sense of convergence that the  sequence $\{x_{k}\}$ generated by \eqref{SA iteration}  actually converges to the root $\theta$ with probability one (almost surely) under weaker conditions. 

Similarly, in order to solve equation \eqref{stationary eq} of data restoration, we consider the following iteration for multivariate case of $\mathbf{x}$, whose dimension is $p$ :
\begin{equation}
\label{ID iteration}
   \hat{\mathbf{x}}_{k+1}-\hat{\mathbf{x}}_k=a_k(\hat{\mathbf{x}}_{0}-\Tilde{\mathbf{x}}_k), \quad k\geq 0 
\end{equation}

where the scalar weight coefficient $a_{k}$ is appropriately selected. $\hat{\mathbf{x}}_{0}$ is the initial data and $\Tilde{\mathbf{x}}_{k}$ is the restored data given input is $\hat{\mathbf{x}}_{k}$, i.e $\Tilde{\mathbf{x}}_{k}$ follows the distribution $F(\Tilde{\mathbf{x}}|\hat{\mathbf{x}}_{k})$.  Denote the $i$-th coordinate of function $H$ as $H_{i}$ and $\hat{\mathbf{x}}_{0}$ as $\alpha=(\alpha_{1},...,\alpha_{p})$ for the clean notations and consider the following set of assumptions $\textbf{A1}-\textbf{A5}$:

\begin{enumerate}[start=1,label={(\bfseries A\arabic*):}] 
\item $\sum_{k=1}^{\infty}a_{k}=\infty,\sum_{k=1}^{\infty}a_{k}^{2}<\infty$.
\item For each  dimension $i \in \{1,2,...p\}$: $H_{i}(\mathbf{x})<\alpha_{i}$ for $x_{i} <\mathbf{x}^{*}_{i}$ and $H_{i}(\mathbf{x})>\alpha_{i}$ for $x_{i} >\mathbf{x}^{*}_{i}$ and $H_{i}(\mathbf{x}^{*})=\alpha_{i}$, where $\mathbf{x}=(x_{1},...,x_{p})\in \mathbb{R}^{d}$ and $\mathbf{x}^{*}=(x^{*}_{1},...,x^{*}_{p})\in \mathbb{R}^{d}$
\item For every $\delta >0$ and dimension $i \in \{1,2,...p\}$, function $H_{i}(\mathbf{x})$ is strictly increasing for each coordinate $j$ w.r.t $\mathbf{x}$, if $|x_{j}-\mathbf{x}^{*}_{j}|<\delta$ where $\mathbf{x}=(x_{1},...,x_{p})\in \mathbb{R}^{d}$ and $\mathbf{x}^{*}=(x^{*}_{1},...,x^{*}_{p})\in \mathbb{R}^{d}$
\item For every $\epsilon >0$ and coordinate $i \in \{1,2,...p\}$ \[\inf_{||\mathbf{x}-\mathbf{x}^{*}||>\epsilon}||H_{i}(\mathbf{x})-\alpha_{i}||>0\] where $||\cdot||$ is $L_{2}$ norm in our setting.
\item  $e_{k}=E\left[E\left[||\Tilde{\mathbf{x}}_{k}-\alpha||^{2}|\hat{\mathbf{x}}_{k}\right]\right] \leq h^{2}
            < \infty$  for each integer $k\geq 0$,
\end{enumerate}

Theorem \ref{convergence DKGM} gives a convergence result of iteration process in \eqref{ID iteration} based on assumptions  $\textbf{A1}-\textbf{A5}$. The proof is a multivariate version of the argument given by \cite{Stochastic_Approximation}. See Appendix \ref{proof of theorem}

\begin{theorem}
\label{convergence DKGM}
   Under assumptions $\textbf{A1}-\textbf{A5}$, the sequence of vector $\{\hat{\mathbf{x}}_{k}\}$ generated by iteration \eqref{ID iteration} converges to  $\mathbf{x}^{*}$ in probability,  which is the solution of  the inverse problem $ \hat{\mathbf{x}}_{0}=H(\mathbf{x}^{*})$.
\end{theorem}

Assumptions $\textbf{A2}-\textbf{A4}$ control the local performance of the map $H$, which implicitly require the decent quality of base model. Indeed, in the experiments, we found that a standard U-Net structure is able to generate good quality of debiased images. The uniform boundedness of the difference between iterated restored data and ideal debiased data in assumption $\textbf{A5}$ can be satisfied as long as the restoration process is stable. 

In section \ref{Bias reduction}, we see that the solution of \eqref{expectation sta eq} is unbiased once the reconstruction map $H$ is linear. In addition, the contracting structure of U-Net can be used to approximate the identity map between original data and debiased data. As a consequence, the learned restoration process is expected to be close to a linear map. This observation makes the idea of iterative debiasing  promising.

The choice of weight coefficients $a_{k}$ in  assumption $\textbf{A1}$ is flexible.  In next section, we will illustrate the connection of weight coefficients with the SDE version of iteration process \eqref{SA iteration}.

\subsection{SDE version of iteration process \eqref{SA iteration} and a choice of weight coefficients $a_{k}$}
\label{SDE of DKGM}
According to \cite{Stochastic_Approximation}, at $k$-th iteration, we can reconstruct $\Tilde{\mathbf{x}}_{k}$  for $m$ times and  replace $\Tilde{\mathbf{x}}_{k}$ in \eqref{ID iteration} with the arithmetic mean of r reconstructions, which is denoted as $\bar{\Tilde{\mathbf{x}}}_{k,m}$, i.e. the iteration \eqref{SA iteration} becomes:
\begin{equation}
\label{DKGM iteration multiple r}
   \hat{\mathbf{x}}_{k+1}-\hat{\mathbf{x}}_k=a_k(\hat{\mathbf{x}}_{0}-\bar{\Tilde{\mathbf{x}}}_{k,m}), \quad k\geq 0 
\end{equation}
where $\bar{\Tilde{\mathbf{x}}}_{k,m}=\frac{\Tilde{\mathbf{x}}_{k1}+..+\Tilde{\mathbf{x}}_{km}}{m}$ and $\Tilde{\mathbf{x}}_{ki},(i=1,2...m)$ are reconstructed data given input data is $\hat{\mathbf{x}}_{k}$.
If we write $a_{k}=\eta u_{k}$, where $\eta$ is fixed step size and $u_{k}$ is adjustable learning rate, iteration \eqref{DKGM iteration multiple r} can be reformulated as follows:
\begin{equation}
\label{sdedevae}
    \begin{split}
        \hat{\mathbf{x}}_{k+1}&=\hat{\mathbf{x}}_{k}+\eta u_{k}(\hat{\mathbf{x}}_{0}-\bar{\Tilde{\mathbf{x}}}_{k,m})\\
        &=\hat{\mathbf{x}}_{k}+\eta u_{k}(\hat{\mathbf{x}}_{0}-E_{\hat{\mathbf{x}}_{k}}[\Tilde{\mathbf{x}}]+ E_{\hat{\mathbf{x}}_{k}}[\Tilde{\mathbf{x}}]- \bar{\Tilde{\mathbf{x}}}_{k,m} )\\
        &=\hat{\mathbf{x}}_{k}+\eta u_{k}(\hat{\mathbf{x}}_{0} -f(\hat{\mathbf{x}}_{k}))+\eta u_{k}(f(\hat{\mathbf{x}}_{k})-\bar{\Tilde{\mathbf{x}}}_{k,m})
    \end{split}
\end{equation}
\cite{SDEDEVAE} gives a weak approximation SDE of \eqref{sdedevae}:
\begin{equation}
\label{constraintsde}
    d\mathbf{x}_{t}=-u_{t}(f(\mathbf{x}_{t})-\mathbf{x}_{0})dt+u_{t}\sqrt{\eta \Sigma(\mathbf{x}_{t})}dW_{t},\mathbf{x}_{0}=\hat{\mathbf{x}}_{0}
\end{equation}
where  $f(x)=E_{\mathbf{x}}[\Tilde{\mathbf{x}}]$,$\Sigma(\mathbf{x})=\frac{1}{m}Var_{\mathbf{x}}(\Tilde{\mathbf{x}})$ and $W_{t}$ is the white noise process. $u_{t}\in [0,1]$ is the continuous time analogue of
the adjustment factor $u_{k}$ with the usual identification $t=k\eta$ and $m$ represents the number of reconstruction times. It's now natural to consider optimizing weight coefficients through the adjustable learning rate $u_{t}$ under the constraint \eqref{constraintsde}. 

\cite{SDEDEVAE} considered a simpler case which assumes that the variance function $\Sigma(\mathbf{x})=\Sigma$ which is constant all the time. If we further assume the linearity of $f$,i.e. $f(\mathbf{x}_{t})=a\mathbf{x}_{t}$, proposition \ref{optimal learning rate} gives us  a version of optimal adjustment of learning rate $u_{t}$. The proof can be found in \cite{SDEDEVAE}.

\begin{proposition}[\cite{SDEDEVAE}]
\label{optimal learning rate}
Denote  $m_{t}= \frac{1}{2}\mathbb{E}[(\mathbf{x}_{t}-\mathbf{x}_{0})^2] $ as the cost function of $u_{t}$. We assume that the variance function $\Sigma(\mathbf{x})=\Sigma>0$ and  $f(\mathbf{x})=a\mathbf{x}, a\in \mathbb{R}$.  Consider the following optimal control problem for learning rate $u_{t}$ with fixed stopping time $T$:
\begin{equation}
\begin{aligned}
\min_{u:[0,T] \to [0,1]} \quad & m_{T}\\
\textrm{s.t.} \quad &      d\mathbf{x}_{t}=-u_{t}(a\mathbf{x}_{t}-\mathbf{x}_{0})dt+u_{t}\sqrt{\eta \Sigma}dW_{t} \\
\quad & \mathbf{x}_{0}=\hat{\mathbf{x}}_{0}\\
\end{aligned}
\end{equation}
then  the optimal control policy would be
\begin{equation}
    u_{t}^{*}= \begin{cases} 
      1 & a\leq 0 \quad or \quad t\leq t^{*}\\
      \frac{1}{1+a(t-t^{*})} & a > 0 \quad or \quad t> t^{*}
   \end{cases}
\end{equation}
where $t^{*}=\frac{1}{2a}log( \frac{4m_{0}}{\eta\Sigma}-1)$.
\end{proposition}

Few observations based on proposition \ref{optimal learning rate}:
\begin{enumerate}
    \item  The optimal adjustment policy implies that the maximum learning rate should be used in initial phase where the drift term is dominant. In the fluctuation phase, the noise term dominates the process so the optimal policy will use the $\sim 1/t$ annealing schedule to reduce the effect of the fluctuation.

    \item  More surprisingly,we can verify such adjustment policy satisfy the sufficient  condition $\textbf{A1}$ of the convergence of Stochastic approximation in Theorem \ref{convergence DKGM} and \cite{Stochastic_Approximation}. ($L_{1}$ divergent and  $L_{2}$ convergent).
\end{enumerate}

Note that it's unrealistic to find the phase change point $t^{*}$ in practice as it requires the knowledge of  coefficient $a$ in $f(\mathbf{x})$ and the constant variance $\Sigma$. However, the $\sim 1/t$ annealing schedule  provides us a candidate of weight coefficient $a_{k}$. For simplicity and stability, we set $a_{k}=\frac{1}{k}$, which is the harmonic series and satisfies the  assumption $\textbf{A1}$ introduced in section \ref{Stochastic approximation}. Our experiments in section \ref{Experiments}  also indicate that setting  $a_{k}=\frac{1}{k}$ is good enough in practice.
\subsection{Choice of network to approximate restored data $\Tilde{\mathbf{x}}$ given input $\hat{\mathbf{x}}_{k}$  }

Vanilla denoising autoencoder (DAE) is a natural choice as it keeps the same input and output size. However, the vanilla DAE has difficulty in recovering original images with simple model structure such as CNN, which is likely to violate the linearity assumption in bias reduction. Another choice is U-Net \cite{DDPM}. Its contracting path (encoder) can fuse the local information with the global feature through skip connections. In that sense, we conjecture that U-Net is better suitable to the assumptions in section \ref{Stochastic approximation}. We then propose the corresponding debiasing structure in section \ref{Iterative Debiasing}. 

\section{Debiasing Kernel Based Generative Models}
\label{Iterative Debiasing}

In this part, we propose a novel two-stage generative model named Debiasing Kernel Based Generative Model (DKGM) based on the idea of KDE and stochastic approximation introduced in section \ref{preliminary}. In section \ref{kernel generative model}, we introduce the first stage of DKGM aiming to build a kernel based generative model , which can be trained as the same in denoising model. Since the initial image generated is blurry (See Fig. \ref{s1-cifar10}) due to the noise added into the input, we then develop a novel iterative debiasing method in section \ref{Iterative debiasing} to enhance the quality of initial generation, which is the second stage of DKGM. The main difference between DKGM and diffusion models is that DKGM directly reduces the bias of initial generated images  while diffusion models requires artificial noises added in the original data. 
\captionsetup[subfigure]{labelformat=simple, labelsep=none}
\renewcommand{\thesubfigure}{(\alph{subfigure})  }
\begin{figure*}
    \subfloat[Transformed data]{\label{Transformed} \includegraphics[width=0.32\columnwidth,height=3cm]{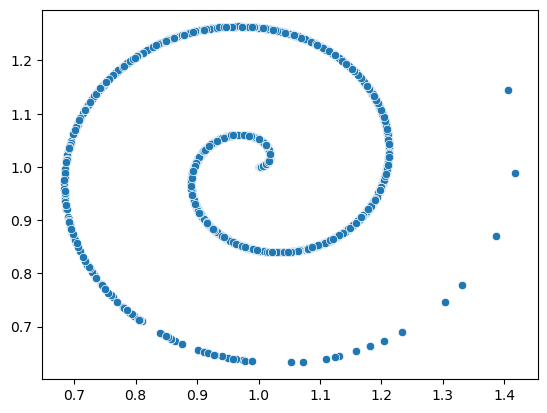}}
\hfill
\subfloat[$k=0$]{\label{k0} \includegraphics[width=0.32\columnwidth,height=3cm]{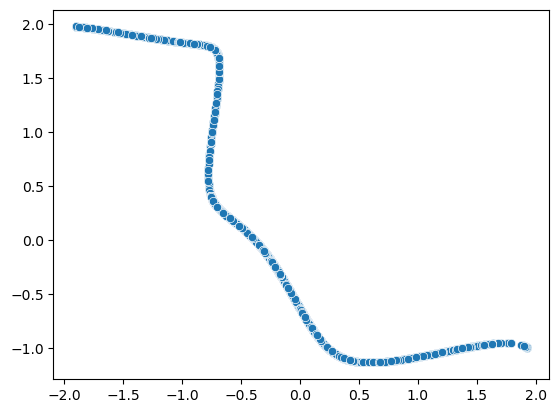}}%
\hfill
\subfloat[$k=1$]{\label{k1} \includegraphics[width=0.32\columnwidth,height=3cm]{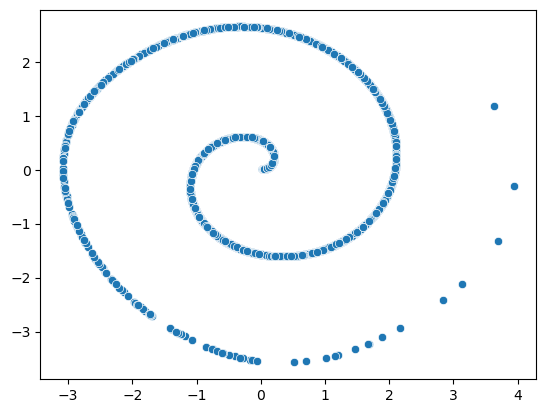}}%
\hfill
\subfloat[$k=5$]{\label{k2} \includegraphics[width=0.32\columnwidth,height=3cm]{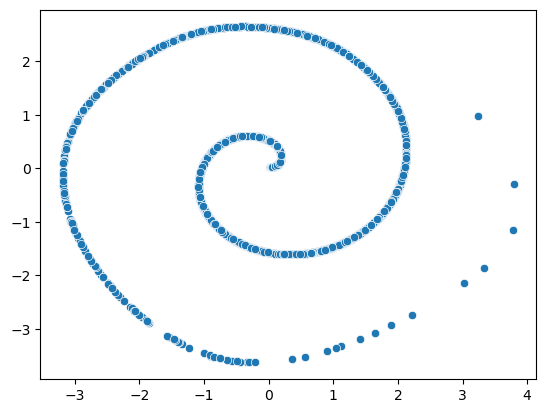}}%
\hfill
\subfloat[$k=10$]{\label{k3} \includegraphics[width=0.32\columnwidth,height=3cm]{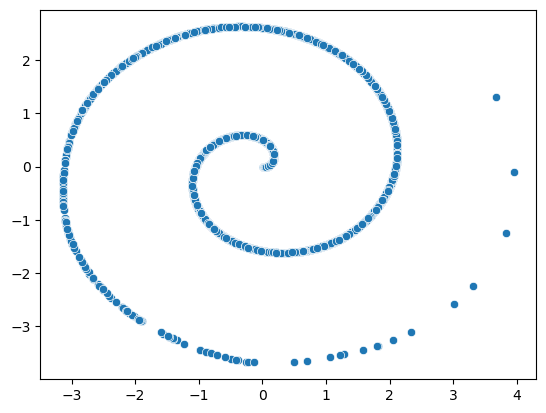}}%
\hfill
\subfloat[Ground Truth]{\label{GroundTruth} \includegraphics[width=0.32\columnwidth,height=3cm]{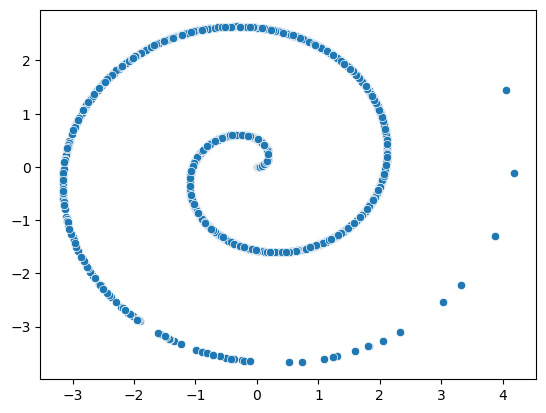}}%
\caption{The Stage 2 DKGM trained on 1-d swiss roll data. The leftmost subplot (a)  is the transformed data, which is the input of stage 2 model. Subplot (f) represents the ground truth.The rest subplots (b)-(d) are reconstructed data corresponding to  different values of $k>0$.  }
\label{swiss roll}
\end{figure*}
\subsection{First stage: Initial generation model}
\label{kernel generative model}

The objective function in first stage is straightforward. To maximize the lower bound of empirical log-likelihood function in \ref{ineq KDE}, we need to maximize the log of KDE for each data point $\mathbf{x}^{(i)}$. For simplicity, we chose gaussian density to be the kernel function in KDE and the target function is reduced to 
\begin{equation}
    \label{target function}
    \sum_{k=1}^{m}||\mathbf{x}^{(i)}-\hat{\mathbf{x}}^{(i)}_{k}||^{2}=\sum_{k=1}^{m}||\mathbf{x}^{(i)}-f_{\mathbf{\theta}}(\mathbf{x}^{(i)}+\mathbf{\epsilon}^{(i)}_{k})||^{2}
\end{equation}

where the map $f_{\mathbf{\theta}}$ is modeled by a neural network and the noise $\mathbf{\epsilon_{k}}$ follows a certain distribution with bandwidth $\mathbf{h}$. The geometric interpretation of the target function \eqref{target function} is, we want the trained model $f_{\mathbf{\theta}}$ to sample the data in the neighborhood of the original input, which is exactly the intuition of KDE. As we can expect, the quality of sampled data simply through $f_{\mathbf{\theta}}$ won't be good enough due to the over smoothness from KDE. For instance, Figure \ref{s1-cifar10} shows that the generated images from stage one are somehow blurry. Therefore, we introduce a debiasing method in second stage to enhance the quality of generated data in stage 1. In our experiments, we found that a large number of KDE samples $m$ per datapoint will lead to out of memory issues in training and we can actually set $m$ to be 1 as long as model $f_{\mathbf{\theta}}$ is strong enough, e.g. U-Net.

\subsection{Second stage: Iterative debiasing}
\label{Iterative debiasing}

\begin{figure}[htp]
\centering
\includegraphics[width=1.0\columnwidth]{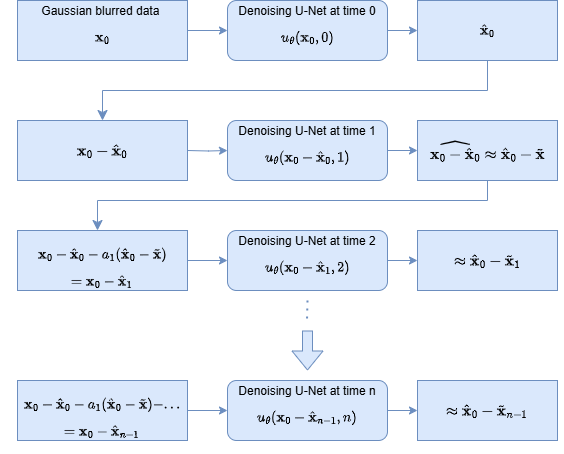}
\caption{The flow chart of second stage  DKGM}
\label{dkgm2}
\end{figure}
 Figure \ref{s1-cifar10} displays the KDE sampled images from stage 1 of DKGM, which are somehow blurry. This phenomenon can be explained by the gaussian noise introduced in each datapoint. By one-dimension Taylor expansion, the function $f_{\theta}$ in \eqref{target function} can be expanded as $f_{\theta}(x^{(i)}+\epsilon)\approx f_{\theta}(x^{(i)})+f'(x^{(i)})\cdot \epsilon$. Then the blurryness comes from the convolution between $f_{\theta}(x^{(i)})$ and $f'(x^{(i)})\cdot \epsilon$, where the second term is equivalent to a gaussian kernel. To reduce the blurriness from the KDE sampling, we introduce the second stage of DKGM with the iteration equation in \eqref{ID iteration}.

In the second stage, we should train another model to reduce the blurriness from first stage. Empirically, we found that it's better to train the debiasing model on Gaussian blurred data rather than the direct KDE samples from the first stage. For Gaussian-blurred data, we meant to perform Gaussian blurring on the image by Gaussian kernel with kernel size (bandwidth) $b$, which is realized by adding a convolution layer in the model. Among our experiments, we found that setting $b$ uniformly distributed during the training epochs will benefit the performance of entire model. The building block of second stage model is still U-Net.
Except for the initial Gaussian blurred data, we iteratively used the output of U-Net to approximate the $t$-th bias term $\hat{\mathbf{x}}_{0}-\Tilde{\mathbf{x}}_{t}$ in iteration \eqref{ID iteration}. See the flow chart in Figure \ref{dkgm2}. There are two strategies of training the iterative restoration described in  Figure \ref{dkgm2}. One is to fully train each U-Net so that we can use it to approximate $\hat{\mathbf{x}}_{0}-\Tilde{\mathbf{x}}_{k}$. However, this strategy requires us to train the U-Nets sequentially which can be time consuming for large number of iteration steps $k$. The other strategy is similar to the implementation in \cite{DDPM}, which trains a time-step embedded U-Net with a single target function. This method is time efficient but has a high demand for memory. In this paper, we decided to employ the second strategy for the sake of time efficiency. Algorithm \ref{stage2 dkgm} summarizes the forward implementation of the second stage of DKGM using the time-step embedded U-Net $u_{\gamma}(\cdot,t)$


\begin{algorithm}[htp]

    \caption{Second stage of DKGM}
        \label{stage2 dkgm}
    \begin{algorithmic}[1]
    
        \Require Gaussian Blurred data $\bar{\mathbf{x}}_{g}$ with bandwidth $b$. Iteration times $n$. Weight coefficients $a_{j}=\frac{1}{j},j=1,...,n$. 
        \State \textbf{Initialization:}
        \State Set $\mathbf{x}_{0}=\bar{\mathbf{x}}_{g}$ and evaluate $\hat{\mathbf{x}}_{0}=u_{\mathbf{\gamma}}(\mathbf{x}_{0},0)$
        \For{$t\in \{1,...,n\}$}
        \State $\hat{\mathbf{x}}_{t}=\hat{\mathbf{x}}_{t-1}+ a_{t}\cdot u_{\mathbf{\gamma}}(\hat{\mathbf{x}}_{0}- \hat{\mathbf{x}}_{t-1},t)$ where we use $u_{\mathbf{\gamma}}(\hat{\mathbf{x}}_{0}- \hat{\mathbf{x}}_{t-1},t)$ to approximate $\mathbf{x}_{0}-\tilde{\mathbf{x}}_{t-1}$
         \EndFor
        \State \textbf{Return:} $\hat{\mathbf{x}}_{n}$
    \end{algorithmic}
\end{algorithm}

Given the output from Algorithm \ref{stage2 dkgm}, we employed the following  comprehensive target function to train the time-step embedded U-Nets simultaneously at $i$-th gaussian-blurred data point $\bar{\mathbf{x}}_{g}^{(i)}$:
\begin{equation}
\label{sample DKGM2}
 \Tilde{\mathcal{L}}(\mathbf{\theta};\hat{\mathbf{x}}^{(i)})= ||\bar{\mathbf{x}}_{g}^{(i)}-\hat{\mathbf{x}}_{n}^{(i)}||^2
\end{equation}
where 
\begin{equation}
\label{generate xk}
    \begin{split}        \hat{\mathbf{x}}_{n}^{(i)}&=\hat{\mathbf{x}}_{0}^{(i)}+\sum_{t=1}^{n}\frac{1}{t}\cdot u_{\mathbf{\gamma}}( \hat{\mathbf{x}}^{(i)}_{0}-\hat{\mathbf{x}}^{(i)}_{t-1},t ),\\
    \hat{\mathbf{x}}^{(i)}_{0}&= u_{\mathbf{\gamma}}(\mathbf{x}_{0}^{(i)},0)=u_{\mathbf{\gamma}}(\bar{\mathbf{x}}^{(i)}_{g},0)
    \end{split}
\end{equation}

The subscript $\mathbf{\gamma}$ represents learnable parameters  in time-step embedded U-Net $u_{\gamma}(\cdot,t)$. With $f_{\theta}$ and $u_{\gamma}(\cdot,t)$ trained, we can use the output of stage 1 KDE sampling as the input of trained stage 2. The detailed  procedure is summarized in Algorithm \ref{sampling dkgm}. 

\begin{algorithm}[H]

    \caption{Sampling with noise level $\alpha$}
        \label{sampling dkgm}
    \begin{algorithmic}[1]  
        \Require $\epsilon \sim N(\mathbf{0},I),\mathbf{x}\sim p_{data},f_{\theta},u_{\gamma}(\cdot,t),\alpha$.
        \State \textbf{Stage 1: KDE sampling}       
        \State $\hat{\mathbf{x}}_{kde}=f_{\theta}(\mathbf{x}+\alpha\epsilon)$
        \State \textbf{Stage 2: Iterative debiasing}
       \State  $\hat{\mathbf{x}}_{0}=u_{\mathbf{\gamma}}(\hat{\mathbf{x}}_{kde},0)$
        \For{$t\in \{1,...,n\}$}
        \State $\hat{\mathbf{x}}_{t}=\hat{\mathbf{x}}_{t-1}+ a_{t}\cdot u_{\mathbf{\gamma}}(\hat{\mathbf{x}}_{kde}- \hat{\mathbf{x}}_{t-1},t)$ where we use $u_{\mathbf{\gamma}}(\hat{\mathbf{x}}_{kde}- \hat{\mathbf{x}}_{t-1},t)$ to approximate $\hat{\mathbf{x}}_{kde}-\tilde{\mathbf{x}}_{t-1}$
         \EndFor
        \State \textbf{Return:} $\hat{\mathbf{x}}_{n}$     
    \end{algorithmic}
\end{algorithm}
\section{Experiments}
\label{Experiments}

In this section, a toy example will be provided to demonstrate the effect of debiasing steps proposed in section \ref{Iterative debiasing}.  We can see increasing the number of bias iterations  indeed benefits the restoration. After that, we compare the sample quality of  DKGM with other state-of-art algorithms in few real datasets. The inception scores and FID scores are reported in Table \ref{sota-table}.   Inception\_v3 \cite{Inceptionv3} is employed as default model in both measurements, which is a standard practice in literature. All model architectures and training details are provided in Appendix \ref{Model architecture} and \ref{Datasets and Training details}.

\begin{figure}
\centering
\includegraphics[width=1.0\columnwidth]{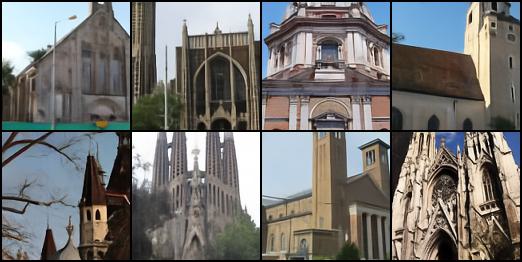}
\caption{LSUN Church samples from DKGM ($\alpha=0.5, b\sim Unif[0.5,1.0]$), FID=4.99}
\label{lsun_dkgm}
\end{figure}

\subsection{Toy example: 1-d Swiss Roll}
\label{toy example}

 \cite{DNF} designs a 1-d swiss roll 
 manifold which is essentially a one dimensional spiral function and vanilla VAEs failed to capture the underlying manifold and density. Additionally,  a one dimensional swiss roll distribution can also be successfully learned by a diffusion model, as illustrated in \cite{Diffusion_original}. Section \ref{Stochastic approximation} implies that stage 2 of DKGM is similar to the diffusion process. Observing how the reconstruction of stage 2 of DKGM evolves with the debiasing steps $k$ in this toy example could give us more insights into the debiasing-type model. For simplicity, the base model of DKGM in this example is vanilla autoencoder. Detailed experiment settings are available in  Appendix \ref{Model architecture}.

Figure \ref{swiss roll} demonstrates that increasing the number of bias iterations in DKGM indeed benefits the reconstructed data. The poor result in vanilla autoencoder reconstruction ($k=0$) is consistent with the results in \cite{DNF} as the latent space learned by the initial vanilla autoencoder is deficient in complicated data manifold and density. The reconstructed manifold from transformed data is closer to the ground truth as $k$ increases. More interestingly, unlike carefully designed transformation functions in normalizing flow models \cite{normalizingflow}, the debiasing part of DKGM shows the potential of learning density on non-euclidean manifolds.

Theoretically, setting large $k$ is always beneficial to the debiased reconstruction, as implied by Theorem \ref{convergence DKGM}.  While in practice, larger $k$ could lead to heavier computation cost and overfitting. How to choose optimal $k$ relies on the complexity of base models and the data. We leave this as the future research.

\begin{table}
\caption{Baseline results on Unconditional CIFAR10 datasets.}
\label{sota-table}
\begin{center}

\begin{tabular}{ccc}
\toprule
Model & IS(std) & FID \\
\midrule
DDPM($L_{simple}$) \cite{DDPM} & 9.46(0.11)& 3.17\\

EBM  \cite{EBM}
 &  6.78(-)& 38.2   \\

NCSN  \cite{NCSN}
 &  8.87(0.12)& 25.32  \\

SNGAN \cite{SNGAN}
&   8.22(0.02)&21.7\\

SNGAN-DDLS   \cite{SNGAN-DDLS}&9.09(0.1)& 15.42  \\

StyleGAN2 + ADA (v1)  \cite{StyleGAN2+ADA} & 9.74(0.05)&3.26 \\

\textbf{DKGM} ($\alpha=0.5, b\sim  Unif[0.5,1]$)
 &  \textbf{8.62(0.18)}&  \textbf{6.42} \\
\bottomrule
\end{tabular}
\end{center}

\end{table}

\subsection{Real datasets}
\label{real datasets}

Table \ref{sota-table} reports Inception scores(IS), FID scores on CIFAR10. We can see our DKGM model achieves comparable Insecption scores and FID scores to many models , including the DDPM.  The FID score is based on the training set, which is standard practice\cite{DDPM}. We also achieved  4.99 FID score in LSUN (Church) 128$\times$ 128 and 11.87 FID score in CelebA 128$\times$ 128. See Fig \ref{lsun_dkgm} for generated sample quality and Appendix \ref{more examples} for more information. The detailed information for training parameters  are attached in Appendix \ref{Datasets and Training details}. 


\subsubsection{Enhancement from Stage 2 of DKGM}

\begin{figure}[htp]
    \subfloat[Stage 1 results \\(FID=44.75,Sharpness=0.0368)]{\label{s1-cifar10} \includegraphics[width=0.49\columnwidth]{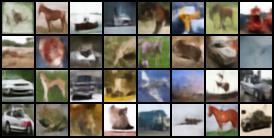}}
\subfloat[Stage 2 results \\(FID=15.76,Sharpness=0.063)]{\label{s2-cifar10} \includegraphics[width=0.49\columnwidth]{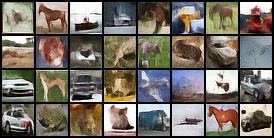}}%
\caption{Randomly generated samples on unconditional CIFAR10 through two stages in DKGM  (a) Generated samples from the model trained in Stage 1 with noise level $\alpha=1.0$ (b) Enhanced samples after stage 2 of DKGM with $b\sim Unif(0.8,1.2)$ }
\label{Example_cifar10}
\end{figure}

Figure \ref{Example_cifar10} provides the generated images on unconditional CIFAR10 using DKGM. Comparing the samples in  Figure \ref{s1-cifar10}  and  Figure \ref{s2-cifar10}, we can see that the debiasing model in stage 2 of DKGM  empirically improves the quality and sharpness of samples generated in stage 1, as verified by FID scores and sharpness. 

\subsubsection{Noise level $\alpha$ in DKGM}
\label{Noise level}
\renewcommand{\thesubfigure}{(\alph{subfigure})  }
\begin{figure}
    \subfloat[Sample diversity $(\alpha=0.5)$ ]{\label{s2-celeba} \includegraphics[width=0.48\columnwidth]{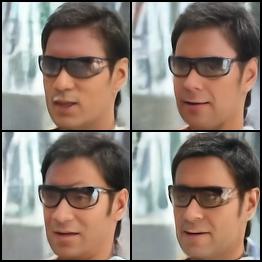}}
\subfloat[Sample diversity $(\alpha=1.0)$ ]{\label{s2-celeba} \includegraphics[width=0.48\columnwidth]{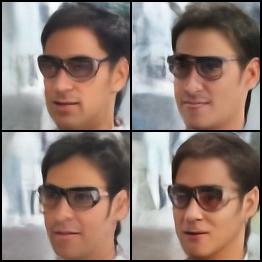}}%
\caption{Randomly generated samples on  CelebA 128$\times$128 with same input under different noise levels of DKGM.  (a) Generated samples from  DKGM trained with noise level $\alpha=0.5$ (b) Generated samples from  DKGM trained with noise level $\alpha=1.0$}
\label{Example_celebA128}
\end{figure}

To see the effect of noise level in stage 1, we provide KDE samples on CelebA 128×128 with same input under different noise levels. When the noise level is low, we see that samples share high-level attribute while when the noise level is high, the samples are more diversed (e.g. hair cut styles are clearly different for samples in Figure \ref{s2-celeba}, which is anticipated according to the idea of KDE.

\subsubsection{Effect of gaussian blurring kernel bandwidth $b$}
\label{bandwidth}
In this section, we performed extra experiments to see how the choice of Gaussian blurring kernel bandwidth would affect the FID, Inception score and sharpness of DKGM on CIFRA10 dataset. The results displayed in Table \ref{bandwidth-table} reveal that tuning bandwidth $b$ can be beneficial and the uniformly distributed bandwidth $b$ with the appropriate range performs better in practice.

\begin{table}[htp]
\caption{Effect of gaussian blurring kernel bandwidth on CIFRA10 $(\alpha=0.5)$}
\label{bandwidth-table}
\vskip -0.1in
\begin{center}
\begin{small}
\begin{sc}
\begin{tabular}{lccc}
\toprule
Bandwidth $b$ & FID & IS(std)& Sharpness  \\
\midrule
$b=0.8$ & 26.55 &8.99(0.09) &0.1406\\
$b=0.6$& 7.65 & 8.72(0.13) &0.0642\\
$b=0.5$   &  7.94&8.27(0.10) & 0.0492\\
$b\sim Unif(0.8,1.2)^{*}$  &  19.4&9.1(0.15) &0.085\\
$b\sim Unif(0.5,1.0)$   &  6.42&8.62(0.18) &0.056\\
\bottomrule
\end{tabular}
\end{sc}
\end{small}
\end{center}
\begin{tablenotes}\scriptsize
\item[*]  *The symbol "$\sim$" means that the bandwidth in guassian blurring layer is uniformly distributed during the training.
\end{tablenotes}
\vskip -0.1in
\end{table}
\subsubsection{Ablation studies of Stage 1 model}

We have theoretically and empirically verified the debiasing efficacy of second stage DKGM. Since the trainings of two stages in DKGM are separated, it's natural to ask would the debiasing algorithm still work if we replace the stage 1 model with other generative models which also incorporate Gaussian noises? Table \ref{ablation} illustrates the change of sample quality of DKGM with different Stage 1(S1) models on CIFAR10 and the results indicate that stage 2 of DKGM performs best when connected with KDE and it indeed benefits the sample quality more or less, except for the FID of DDPM. One possible reason is that the backward diffusion process focuses more on distributional characteristics and less on details. However, stage 2 of DKGM improved sharpness in general, indicating it is able to fill in the "gaps" in terms of sharpness. The stage 2 DKGM trained with $b\sim Unif[0.5,1]$ is used for all  S1 models listed in Table \ref{ablation}. FID and Sharpness are evaluated on testing set. The S1 model implementations all follow  the standard practice. (See more details in Appendix \ref{Ablation study})

\begin{table}[htp]

\caption{Change of sample quality of DKGM with different S1 models on Unconditional CIFAR10}
\label{ablation}
\begin{center}
\begin{tabular}{lcc}
\toprule
S1 model&  $\Delta$FID(\%)* &$\Delta$Sharpeness(\%)  \\
\midrule
VAE & -9\% & 4\%\\
DCGAN& -10\%& 45\%\\
DDPM & 24\% & 55\%\\
\textbf{Ours KDE model ($\alpha=0.5$)} &-64\%& 36\% \\
\bottomrule
\end{tabular}
\end{center}
\begin{tablenotes}\scriptsize
\item[*]  *$\Delta$ FID(\%) is the increase/decrease percentage of FID for DKGM with new S1 models with respect to S1 models. Similar definition for $\Delta$Sharpness(\%)
\end{tablenotes}

\end{table}

\section{Limitation and discussion}
\label{discussion}

\subsection{Limitations of DKGM}

\textbf{Time efficiency of training stage 2 of DKGM}

As we mentioned in section \ref{Iterative debiasing}, the current implementation of stage 2 DKGM training can be time consuming if we needs more iteration steps $k$ in debiasing step\footnote{In most of experiments, we set $k=4$, which is good enough.}. One possible reason is the choice of base model in the iteration. Our experiments indicate that U-Net is indeed a good choice while the time efficiency of deliberate network would be the sacrifice. How to design a more efficient network structure (e.g. CNN) is one of our future works.

\textbf{Trade off between sample quality and sample diversity}

In sections \ref{Noise level} and \ref{bandwidth}, we have seen that the noise level in stage 1 of DKGM controls the sample diversity and the bandwidth in stage 2 affects the sample quality. Table \ref{bandwidth-table}  implies that there might be an optimal bandwidth given each noise level. How to incorporate noise level and blurring bandwidth is an interesting open question. On the other hand, it's possible to embed these two parameters into the training of DKGM and we will leave it as future research.

\subsection{Connection between score-based models}

Recall the SDE \eqref{constraintsde} in section \ref{SDE of DKGM}, a stochastic version of iterative debiasing in stage 2 of DKGM:
\[
    d\mathbf{x}_{t}=-u_{t}(f(\mathbf{x}_{t})-\mathbf{x}_{0})dt+u_{t}\sqrt{\eta \Sigma(\mathbf{x}_{t})}dW_{t},\quad \mathbf{x}_{0}=\hat{\mathbf{x}}_{0}
\]
where  $f(\mathbf{x}_{t})=E[\Tilde{\mathbf{x}}_{t}|\mathbf{x}_{0}]$, $\Sigma(\mathbf{x})=\frac{1}{m}Var_{\mathbf{x}}(\Tilde{\mathbf{x}})$ ,$W_{t}$ is the white noise process and $\mathbf{x}_{0}$ is the initial data. With Tweedie's formula \cite{Efron01122011}, we have $E[\Tilde{\mathbf{x}}_{t}|\mathbf{x}_{0}]\approx \mathbf{x}_{0}+\sigma^{2}\delta \cdot \nabla_{\hat{\mathbf{x}}_{0}}\text{log} p_{\delta}(\mathbf{x}_{0})$ if $\mathbf{x}_{0}=\Tilde{\mathbf{x}}_{t}+N(0,\sigma^{2}\delta)$. For simplicity, we assume $u_{t}=1$ for $t\in [0,1]$. Then the SDE above becomes
\[
\label{tweedie ie}
    d\mathbf{x}_{t}=-\nabla_{\hat{\mathbf{x}}_{0}} p_{\delta}(\mathbf{x}_{0})dt+\sqrt{\eta \Sigma(\mathbf{x}_{t})}dW_{t}
\]
where $p_{\delta}(\mathbf{x}_{0})$ represents the density of $\mathbf{x}_{0})$ given parameter $\delta$ and the term $\nabla_{\hat{\mathbf{x}}_{0}}\text{log} p_{\delta}(\mathbf{x}_{0})$ is typically called score function of distribution $p_{\delta}(\mathbf{x}_{0})$ in literature \cite{score-based,Song2019SlicedSM,Song2020ScoreBasedGM}. To this sense, DKGM is somehow related with score-based models if the posterior expectation term  $f(\mathbf{x}_{t})=E[\Tilde{\mathbf{x}}_{t}|\mathbf{x}_{0}]$ has appropriate forms. This observation further demonstrates the flexibility of DKGM as we don't explicitly assume the functional form of $E[\Tilde{\mathbf{x}}_{t}|\mathbf{x}_{0}]$ at stage 2 of DKGM.

\subsection{Data augmentation}
\label{Downstream task}

\renewcommand{\thesubfigure}{}
\begin{figure}[htp]
\subfloat[$\mathbf{x}_{0}$]{\label{cifar10eg1} \includegraphics[width=0.105\columnwidth]{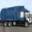}}%
\subfloat[KDE samples $\mathbf{x}_{k}$ from DKGM]{\label{cifar10keg1} \includegraphics[width=0.84\columnwidth]{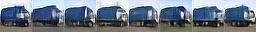}}%
\hfill
\subfloat[$\mathbf{x}_{0}$]{\label{cifar10eg2} \includegraphics[width=0.105\columnwidth]{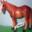}}%
\subfloat[KDE samples $\mathbf{x}_{k}$ from DKGM]{\label{cifar10keg2} \includegraphics[width=0.84\columnwidth]{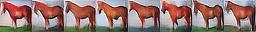}}%
\caption{KDE samples on Cifar10 under the same input $\mathbf{x}_{0}$ with DKGM trained under noise level $\alpha=0.5$. The leftmost images are the input of DKGM and the rest of images are generated images from DKGM.}
\label{KDE_cifar10}
\end{figure}

As shown in Figure \ref{KDE_cifar10}, by DKGM we can generate high-quality similar images in the neighborhood of the input image. One possible downstream application would be the data augmentation in image classification tasks. The generation step in DKGM is essentially a type of KDE sampling, which generates similar images in the neighborhood of the input image. The edge is that if we can sample more representative images for certain labels, the trained classifier can see more informative examples for underrepresented labels, which is especially useful for imbalanced datasets such as cancer detection. In that way, we can augment the original training set with DKGM to boost  the classifier performance.


\section*{Impact Statement}

This paper aims to propose a novel generative model. The theoretical results should not have negative societal impacts. One possible negative impact resulting from DKGM might be the misuse  in producing fake images which may lead to security issues in some face recognition based systems. Few mitigation strategies: (1) gate the release of models for commercial use;(2) add a mechanism for monitoring fake images generated by models such as the discriminator in GAN. All benchmark datasets used in this paper are public and well known to the machine learning community. 


\bibliography{reference}
\bibliographystyle{plain}

\newpage
\appendix
\onecolumn
\section{Proofs}

\subsection{Proof of Theorem \ref{convergence DKGM}}
\label{proof of theorem}

Let $b_{k}=E||\hat{\mathbf{x}}_{k}-\mathbf{x}^{*}||^2$.  We can then express $b_{n+1}$ as following:
\begin{equation} 
\label{bk iteration}
\begin{split}
b_{k+1}&=E||\hat{\mathbf{x}}_{k+1}-\mathbf{x}^{*}||_{2}^{2}=E[E(||\hat{\mathbf{x}}_{k+1}-\mathbf{x}^{*}||_{2}^{2}|\hat{\mathbf{x}}_{k}]\\
&=E[E[(\hat{\mathbf{x}}_{k+1}-\mathbf{x}^{*})^{T}(\hat{\mathbf{x}}_{k+1}-\mathbf{x}^{*})|\mathbf{x}^{*}]]\\
&=E[E[(\hat{\mathbf{x}}_k-\mathbf{x}^{*}+a_k(\mathbf{\alpha}-\Tilde{\mathbf{x}}_k))^{T}(\hat{\mathbf{x}}_k-\mathbf{x}^{*}+a_k(\mathbf{\alpha}-\Tilde{\mathbf{x}}_k)|\hat{\mathbf{x}}_{k}]]\\
&=E[E[(\hat{\mathbf{x}}_k-\mathbf{x}^{*})^{T}(\hat{\mathbf{x}}_k-\mathbf{x}^{*})+2a_{k}(\hat{\mathbf{x}}_k-\mathbf{x}^{*})^{T}(\mathbf{\alpha}-\Tilde{\mathbf{x}}_{k})+a_{k}^{2}(\mathbf{\alpha}-\Tilde{\mathbf{x}}_{k})^{T}(\mathbf{\alpha}-\Tilde{\mathbf{x}}_{k})|\hat{\mathbf{x}}_{k}]]\\
&=b_{k}-2a_{k}E[(\hat{\mathbf{x}}_k-\mathbf{x}^{*})^{T}(H(\hat{\mathbf{x}_{k}})-\mathbf{\alpha})]+a_{k}^{2}E[E[||(\mathbf{\alpha}-\Tilde{\mathbf{x}}_{k})||_{2}^{2}|\hat{\mathbf{x}}_{k}]]
\end{split}
\end{equation}

Note that $E[(\hat{\mathbf{x}}_k-\mathbf{x}^{*})^{T}(H(\hat{\mathbf{x}_{k}})-\mathbf{\alpha}))]$ is a scalar, we can simplify this expression by trace operation:
\begin{align*}
E[(\hat{\mathbf{x}}_k-\mathbf{x}^{*})^{T}(H(\hat{\mathbf{x}_{k}})-\mathbf{\alpha}))] &= E[tr\{(\hat{\mathbf{x}}_k-\mathbf{x}^{*})^{T}(H(\hat{\mathbf{x}_{k}})-\mathbf{\alpha}))\}]\\
&=E[tr\{(H(\hat{\mathbf{x}_{k}})-\mathbf{\alpha}))(\hat{\mathbf{x}}_k-\mathbf{x}^{*})^{T}\}]\\
&=tr\{E[(H(\hat{\mathbf{x}_{k}})-\mathbf{\alpha}))(\hat{\mathbf{x}}_k-\mathbf{x}^{*})^{T}]\}\\
&=tr\{B_{k}\}
\end{align*}

where $B_{k}=E[(H(\hat{\mathbf{x}_{k}})-\mathbf{\alpha}))(\hat{\mathbf{x}}_k-\mathbf{x}^{*})^{T}]$ and the third equation comes from the linearity of expectation and trace operation.

Now the equation \eqref{bk iteration} becomes:

\begin{equation}
    b_{k+1}=b_{k}-2a_{k}tr\{B_{k}\}+a_{k}^{2}E[E[||(\mathbf{\alpha}-\Tilde{\mathbf{x}}_{k})||_{2}^{2}|\hat{\mathbf{x}}_{k}]]
\end{equation}

Summing up first $k+1$ equations, we obtain:

\[
b_{k+1}=b_{1}-2\sum_{i=0}^{k}a_{i}tr\{B_{i}\}+\sum_{i=0}^{n}a_{i}^{2}E[E[||(\mathbf{\alpha}-\Tilde{\mathbf{x}}_{i})||_{2}^{2}|\hat{\mathbf{x}}_{i}]]
\]

Since $b_{k+1} \geq 0$ by definition, we have

\begin{equation}
    \begin{split}
        \sum_{i=0}^{k}a_{i}tr\{B_{i}\}&\leq \frac{1}{2}\bigg(b_{0}+\sum_{i=0}^{n}a_{i}^{2}E[E[||(\mathbf{\alpha}-\Tilde{\mathbf{x}}_{i})||_{2}^{2}|\hat{\mathbf{x}}_{i}]]\bigg)\\
        &\leq \frac{1}{2}\left( b_{0}+\sum_{i=0}^{k}a_{i}^{2}h^{2} \right)\quad \text{(Assumption \textbf{A5})}\\
    \end{split}
\end{equation}

By Assumption \textbf{A1}, we can conclude that

\begin{equation}
    \label{akBk bound }
\sum_{i=1}^{\infty}a_{i}tr\{B_{i}\} \leq  \frac{1}{2}\bigg(b_{1}+\sum_{i=1}^{\infty}a_{i}^{2}h^{2}\bigg) < \infty
\end{equation}

So $\sum_{i=1}^{\infty}a_{i}tr\{B_{i}\}$ exists and \[ \lim_{k \to \infty}a_{k} tr\{B_{k}\}  = 0 \].

We claim that the sequence $\hat{\mathbf{x}}_{k}$ converges to $\mathbf{x}^{*}$ in probability. Otherwise, there exists an subsequence $\{k_{j},j=1,2,...\}$ of sequence $\{k\}$ and positive number $\epsilon,\eta$ such that 

\[
P\{||\hat{\mathbf{x}}_{k_{j}}-\mathbf{x}^{*}||>\eta\} >\epsilon
\]
for all $k_{j}$.
On the other hand, for all $k_{j}$ we  have

\begin{align*}
tr(B_{k_{j}})&= tr\{E[(H(\hat{\mathbf{x}}_{k_{j}})-\mathbf{\alpha}))(\hat{\mathbf{x}}_{k_{j}}-\mathbf{x}^{*})^{T}]\}\\
&=E[tr\{(H(\hat{\mathbf{x}}_{k_{j}})-\mathbf{\alpha}))(\hat{\mathbf{x}}_{k_{j}}-\mathbf{x}^{*})^{T}\}]\\
&=E[(\hat{\mathbf{x}}_{k_{j}}-\mathbf{x}^{*})^{T}(H(\hat{\mathbf{x}}_{k_{j}})-\mathbf{\alpha}))]\\
&=E\bigg[\sum_{i=1}^{p}(\hat{\mathbf{x}}_{i,k_{j}}-\mathbf{x}^{*}_{i})(H_{i}(\hat{\mathbf{x}}_{k_{j}})-\alpha_{i})\bigg]\\
&=E\bigg[\sum_{i=1}^{p}|\hat{\mathbf{x}}_{i,k_{j}}-\mathbf{x}^{*}_{i}||H_{i}(\hat{\mathbf{x}}_{k_{j}})-\alpha_{i}|\bigg]\\
& \geq \epsilon\eta \inf_{||\mathbf{x}-\mathbf{x}^{*}||>\epsilon}||H_{i}(\mathbf{x})-\alpha_{i}||>0
\end{align*}
for some $i \in \{1,2,...p\}$.
The last three lines come from $\textbf{A2}-\textbf{A4}$

It follows that

\[
tr(B_{k_{j}}) >0 
\]

for all $k_{j}$, which contradicts the inequality \eqref{akBk bound }. We then proves the claim.

\section{Toy example: 1-d Swiss Roll}
\label{Toy example}

Similar to the construction in \ref{toy example}, we consider a similar 1 dimensional manifold embedded in $R^{2}$, 
a thin spiral. Suppose the latent variable $u$ follows an
exponential distribution with rate 1, i.e. the density is $f(u)=e^{-u}$, and we generate the spiral through function 

\[
g(u) =\frac{\alpha \sqrt{u}}{3}(cos(\alpha\sqrt{u}),sin(\alpha\sqrt{u}))^{T},\quad \alpha=\frac{4\pi}{3}
\]

See ground truth plot in Figure \ref{GroundTruth}. To validate the reconstruction performance of stage 2 DKGM, we perform a simple linear transformation of the ground truth, i.e the transformed spiral has form of 
\[
\tilde{g}(u)=0.1g(u)+1
\]

whose plot is illustrated in Figure \ref{Transformed} for transformed data.




\section{Model architecture}
\label{Model architecture}
\subsection{ 1-d Swiss Roll}

Since this is a simple dataset, We employed simple auto-encoders as base models in DKGM. The encoder structure involves two layer of linear mapping composed with tanh sigmoid function . In the last fully connect layer, we set the latent dimensions $d_{z}=1$ 

\textbf{Encoder:}

\begin{equation*}
    \begin{split}
        x\in \mathbb{R}^{2} &\to \text{Linear layer $W^{2\times 4}$} \to \text{Tanh} \\
        &\to \text{Linear layer $W^{4\times 8}$} \to \text{Tanh} \\
        &\to \text{Fully connected ($8 \times d_{z}$) for each parameters}
    \end{split}
\end{equation*}

\textbf{Decoder:}

\begin{equation*}
    \begin{split}
        z\in \mathbb{R}^{d_{z}} &\to \text{Fully connected ($ d_{z}\times 4  $)}\to \text{Tanh}\\      
        &\to \text{Linear layer $W^{4\times 8}$} \to \text{Tanh} \\
        &\to \text{Linear layer $W^{8\times2}$} \to \in \mathbb{R}^{2}
    \end{split}
\end{equation*}

\subsection{CIFAR-10}

The U-Net implementation for CIFAR-10 follows the work in \cite{DDPM}. More specifically, for the time embedding, all parameters are shared across time, which is specified
to the network using the Transformer sinusoidal position embedding \cite{DDPM}. Self-attentions are employed at
the 16 $\times$ 16 feature map resolution.

\subsection{CelebA 128$\times$ 128 and LSUN (Church) 128$\times$ 128}

For CelebA 128$\times$128 and LSUN (Church) 128$\times$ 128, the U-Net is adjusted to the input size of 128$\times$128 and the self-attentions blocks are removed to reduce the model size.



\section{Datasets and Training details}
\label{Datasets and Training details}
We list details for each benchmark dataset in following table

\begin{center}
\begin{tabular}{ |l|c|c|c| }
\hline
Datasets&\# Training samples& \# Hold-out samples & Original image size\\
\hline
CIFAR-10&50000&10000&32*32\\
\hline
CelebA 128$\times$128&162770&19867 &178*218\\
\hline
LSUN (Church) 128$\times$128&126227&- &256*256\\
\hline
\end{tabular}
\end{center}

Note that for CIFAR-10, we used default splittings of training sets and testing sets provided in Pytorch (torchvision.datasets). For CelebA 128$\times$128, we used default validation set as hold-out samples. As to LSUN (Church), the FID score is evaluated on the whole training set for the sake of comparison.

For the training details, we used same training parameters for all algorithms and datasets, as described in following table

\begin{center}
\begin{tabular}{ |l|c| }
\hline
DKGM iteration times $k$& 4\\
\hline
Optimizer& Adam with learning rate 3e-4\\
\hline
Batch size& 100 (20 for CelebA and LSUN)\\
\hline
Epochs& 50 (20 for CelebA and LSUN)\\
\hline
\end{tabular}
\end{center}

 Our CIFAR10 model has 35.7 million parameters, and our LSUN and CelebA models have 114 million parameters.  We used NVIDA 4070 GPU (8G) for all experiments. Under the training settings in Table above, the training of whole stage of DKGM for CIFAR10 requires 8 hours (2.5 hours for Stage 1 while 5.5 hours for Stage 2) and the training time for CelebA 128$\times$128 and LSUN (Church) $128\times128$ is nearly 22 hours. (5 hours for Stage 1 training and 17 hours for Stage 2 training.
 
\textbf{Calculation of Sharpness}

We follow the way in \cite{WAE} in calculating the sharpness of an image. For each generated image, we first transformed it into grayscale and  convolved it with the Laplace filter$\begin{pmatrix}
0 & 1 &0\\
1 & -4 &1\\
0 & 1 &0\\
\end{pmatrix}$, computed the variance of the resulting activations and took the average of all variances. The resulting number is denoted as sharpness (larger is better). The blurrier image will have less edges. As a result, the variance of activations will be small as most activations will be close to zero. Note that we averaged the sharpness of all reconstructed images from hold-out samples for each dataset.

\section{Ablation studies}
\label{Ablation study}

The implementation of VAE follows a standard encoder-decoder structure. We used fully convolutional architectures with 4 × 4 convolutional filters for both encoder and decoder in  VAE. In encoder, we employed a layer of Adaptive Average pool filter.

\textbf{Encoder $q_{\phi}$:}

\begin{equation*}
    \begin{split}
        x\in \mathbb{R}^{32\times 32} &\to \text{32 Conv, Stride 2} \to \text{BatchNorm} \to \text{ReLU}\\
        &\to \text{64 Conv, Stride 2} \to \text{BatchNorm} \to \text{ReLU}\\
        &\to \text{128 Conv, Stride 2} \to \text{BatchNorm} \to \text{ReLU}\\
        &\to \text{256 Conv, Stride 2} \to \text{BatchNorm} \to \text{ReLU}\\
        &\to \text{AdaptiveAvgPool2d}\\
        &\to \text{Fully connected ($1*1*256 \times d_{z}$) for each parameters}
    \end{split}
\end{equation*}

\textbf{Decoder $p_{\theta}$:}

\begin{equation*}
    \begin{split}
        z\in \mathbb{R}^{d_{z}\times d_{z}} &\to \text{Fully connected ($ d_{z}\times 1*1*256  $)}\\      
        &\to \text{128 ConvTran, Stride 1} \to \text{BatchNorm} \to \text{ReLU}\\
        &\to \text{64 ConvTran, Stride 2} \to \text{BatchNorm} \to \text{ReLU}\\
        &\to \text{32 ConvTran, Stride 2} \to \text{BatchNorm} \to \text{ReLU}\\
        &\to \text{3 ConvTran, Stride 2} \to \text{Sigmoid}
    \end{split}
\end{equation*}

The implementation of DCGAN\cite{DCGAN} follows the popular structure proposed in \cite{DCGAN}. And we still employed $4\times 4$  convolutional filters.

\textbf{Generator $G(z)$:}

\begin{equation*}
    \begin{split}
        z\in \mathbb{R}^{100}    &\to \text{512 ConvTran, Stride 1} \to \text{BatchNorm} \to \text{ReLU}\\
        &\to \text{256 ConvTran, Stride 2} \to \text{BatchNorm} \to \text{ReLU}\\
        &\to \text{128 ConvTran, Stride 2} \to \text{BatchNorm} \to \text{ReLU}\\
        &\to \text{3 ConvTran, Stride 2} \to \text{Tanh}
    \end{split}
\end{equation*}

\textbf{Discriminator $D(x)$:}

\begin{equation*}
    \begin{split}
       x\in \mathbb{R}^{32\times32}    &\to \text{64 Conv, Stride 2} \to \text{BatchNorm} \to \text{ReLU}\\
        &\to \text{128 Conv, Stride 2} \to \text{BatchNorm} \to \text{ReLU}\\
        &\to \text{256 Conv, Stride 2} \to \text{BatchNorm} \to \text{ReLU}\\
        &\to \text{1 Conv, Stride 1} \to \text{Sigmoid}
    \end{split}
\end{equation*}

The implementation of DDPM follows the structures in \cite{DDPM}. All those S1 models are trained with Batch size 32, Adam learning rate 3e-4 and 50 Epochs. 

\textbf{Using VAE as stage 1 model }

\begin{figure}[H]
\centering
\begin{subfigure}{.45\textwidth}
  \centering
  \includegraphics[width=0.9\linewidth]{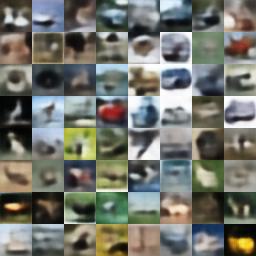}
  \caption{Sample from VAE \\(FID=120.9, Sharpness=0.0364)}
  \label{fig:sfig1}
\end{subfigure}%
\begin{subfigure}{.45\textwidth}
  \centering
  \includegraphics[width=.9\linewidth]{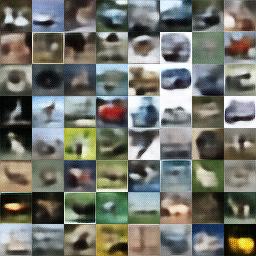}
  \caption{  Bias reduced sample with Stage 2 DKGM with $b\sim Unif(0.5,1.0)$ (FID=109.8, Sharpness=0.0379)}
  \label{fig:sfig2}
\end{subfigure}
\caption{}
\label{fig:fig}
\end{figure}

\textbf{Using DCGAN as stage 1 model }

\begin{figure}[H]
\centering
\begin{subfigure}{.45\textwidth}
  \centering
  \includegraphics[width=0.9\linewidth]{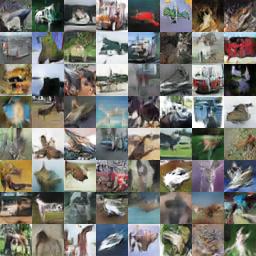}
  \caption{Generated sample DCGAN (FID=15.42, Sharpness=0.0422)}
  \label{fig:sfig1}
\end{subfigure}%
\begin{subfigure}{.45\textwidth}
  \centering
  \includegraphics[width=.9\linewidth]{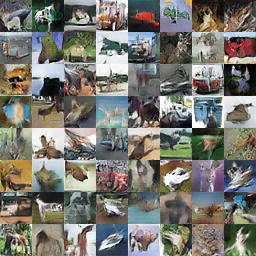}
  \caption{ Bias reduced sample with Stage 2 DKGM with $b\sim Unif[0.5,1.0]$ (FID=13.89, Sharpness=0.0585) }
  \label{fig:sfig2}
\end{subfigure}
\caption{}
\label{fig:fig}
\end{figure}
\newpage
\textbf{Using DDPM as stage 1 model }

\begin{figure}[H]
\centering
\begin{subfigure}{.45\textwidth}
  \centering
  \includegraphics[width=0.9\linewidth]{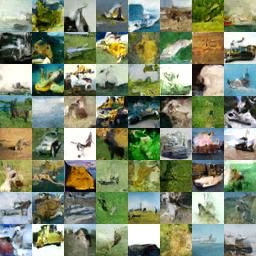}
  \caption{Generated sample DDPM \\(FID=26.45, Sharpness=0.0433)}
  \label{fig:sfig1}
\end{subfigure}%
\begin{subfigure}{.45\textwidth}
  \centering
  \includegraphics[width=.9\linewidth]{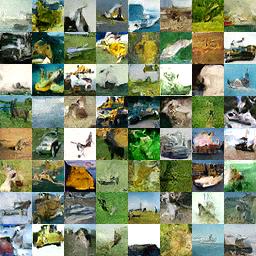}
  \caption{ Bias reduced sample with Stage 2 DKGM with $b\sim Unif[0.5,1.0]$ (FID=35.06, Sharpness=0.0669) }
  \label{fig:sfig2}
\end{subfigure}
\caption{}
\label{fig:fig}
\end{figure}

\newpage
\section{More generated images from DKGM}
\label{more examples}
\subsection{Unconditional CIFAR10}


\begin{figure}[htp]
\centering
\begin{subfigure}{.45\textwidth}
  \centering
  \includegraphics[width=0.9\linewidth]{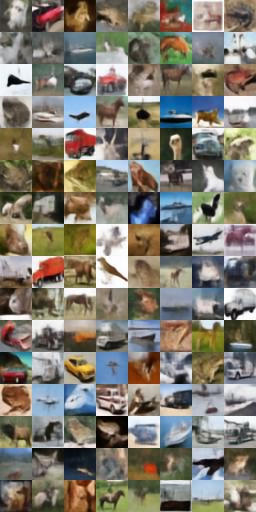}
  \caption{Generated sample from Stage 1 DKGM with $\alpha=1$ (FID=44.75, Sharpness=0.0368)}
  \label{fig:sfig1}
\end{subfigure}%
\begin{subfigure}{.45\textwidth}
  \centering
  \includegraphics[width=0.9\linewidth]{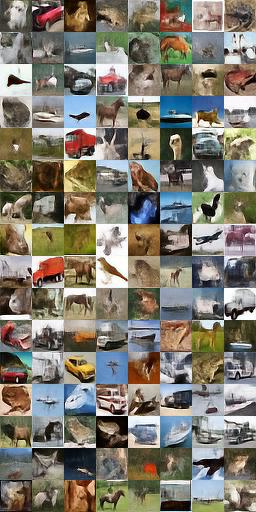}
  \caption{    Bias reduced sample with Stage 2 DKGM with $b\sim Unif[0.8,1.2]$ (FID=15.76, Sharpness=0.063)}
  \label{fig:sfig2}
\end{subfigure}
\caption{}
\label{fig:fig}
\end{figure}
\newpage
\subsection{CelebA $128\times 128$}

\begin{figure}[htp]
\centering
\begin{subfigure}{.5\textwidth}
  \centering
  \includegraphics[width=0.98\columnwidth]{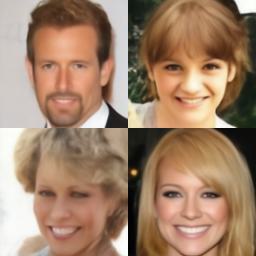}
  \caption{Generated sample from Stage 1 DKGM with $\alpha=0.5$ (FID=15.57, Sharpness=0.0121)}
  \label{fig:sfig1}
\end{subfigure}%
\begin{subfigure}{.5\textwidth}
  \centering
  \includegraphics[width=0.98\columnwidth]{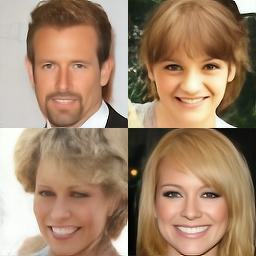}
  \caption{Bias reduced sample with Stage 2 DKGM with $b\sim Unif[0.5,1.0]$(FID=11.87, Sharpness=0.0155)}
  \label{fig:sfig2}
\end{subfigure}
\caption{}
\label{fig:fig}
\end{figure}

\subsection{LSUN Church $128\times 128$}

\begin{figure}[H]
\centering
\begin{subfigure}{.5\textwidth}
  \centering
  \includegraphics[width=0.98\columnwidth]{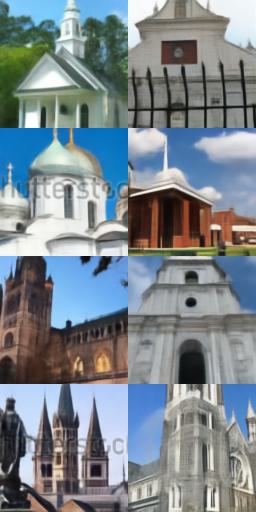}
  \caption{Generated sample from Stage 1 DKGM with $\alpha=1$ (FID=22.10, Sharpness=0.0127)}
  \label{fig:sfig1}
\end{subfigure}%
\begin{subfigure}{.5\textwidth}
  \centering
  \includegraphics[width=0.98\columnwidth]{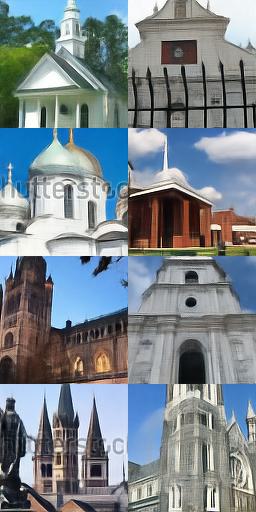}
  \caption{Bias reduced sample with Stage 2 DKGM with $b\sim Unif(0.8,1.2)$(FID=13.93, Sharpness=0.0312)}
  \label{fig:sfig2}
\end{subfigure}
\caption{}
\label{fig:fig}
\end{figure}

\begin{figure}[H]
\centering
\begin{subfigure}{.5\textwidth}
  \centering
  \includegraphics[width=0.98\columnwidth]{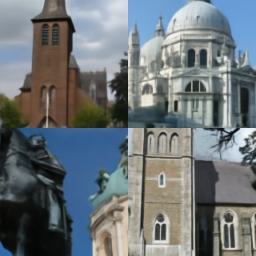}
  \caption{Generated sample from Stage 1 DKGM with $\alpha=0.5$(FID=7.77, Sharpness=0.0151)}
  \label{fig:sfig1}
\end{subfigure}%
\begin{subfigure}{.5\textwidth}
  \centering
  \includegraphics[width=0.98\columnwidth]{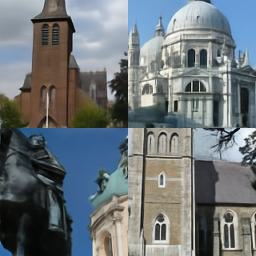}
  \caption{Bias reduced sample with Stage 2 DKGM with $b\sim Unif[0.5,1.0]$(FID=4.99, Sharpness=0.0262)}
  \label{fig:sfig2}
\end{subfigure}
\caption{}
\label{fig:fig}
\end{figure}

\begin{figure}[htp]
\subfloat[$\mathbf{x}_{0}$]{\label{cifar10eg1} \includegraphics[width=0.19\columnwidth]{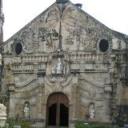}}%
\subfloat[KDE samples $\mathbf{x}_{k}$ from DKGM with $\alpha=1$, $b\sim Unif(0.8,1.2)$ ]{\label{cifar10keg1} \includegraphics[width=0.76\columnwidth]{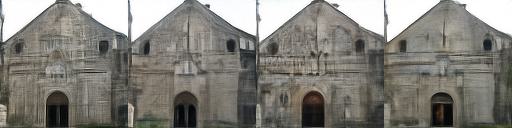}}%
\hfill
\subfloat[$\mathbf{x}_{0}$]{\label{cifar10eg2} \includegraphics[width=0.19\columnwidth]{LSUN_KDE_ori_a1.jpg}}%
\subfloat[KDE samples $\mathbf{x}_{k}$ from DKGM with $\alpha=0.5$, $b\sim Unif(0.5,1.0)$]{\label{cifar10keg2} \includegraphics[width=0.76\columnwidth]{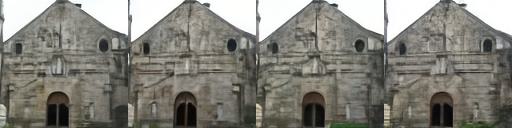}}%
\caption{KDE samples on LSUN Church under the same input $\mathbf{x}_{0}$ with DKGM trained under different noise levels. The leftmost images are the input of DKGM and the rest of images are generated images from DKGM.}
\label{KDE_lsun}
\end{figure}

\end{document}